# WordNet-Based Information Retrieval Using Common Hypernyms and Combined Features

Vuong M. Ngo
HCMC University of Technology
and John von Neumann Institute
VNU-HCM, Viet Nam

Tru H. Cao
HCMC University of Technology
and John von Neumann Institute
VNU-HCM, Viet Nam

Tuan M. V. Le
HCMC University of Technology
and John von Neumann Institute
VNU-HCM, Viet Nam

vuong.cs@gmail.com

## ABSTRACT
Text search based on lexical matching of keywords is not satisfactory due to polysemous and synonymous words. Semantic search that exploits word meanings, in general, improves search performance. In this paper, we survey WordNet-based information retrieval systems, which employ a word sense disambiguation method to process queries and documents. The problem is that in many cases a word has more than one possible direct sense, and picking only one of them may give a wrong sense for the word. Moreover, the previous systems use only word forms to represent word senses and their hypernyms. We propose a novel approach that uses the most specific common hypernym of the remaining undisambiguated multi-senses of a word, as well as combined WordNet features to represent word meanings. Experiments on a benchmark dataset show that, in terms of the MAP measure, our search engine is 17.7% better than the lexical search, and at least 9.4% better than all surveyed search systems using WordNet.

## 1. INTRODUCTION
By today, a large amount of information is shared on the World Wide Web. The performance improvement of rich and huge information exploitation has been challenging research on information retrieval (IR). To overcome the disadvantages of the keyword-based IR models, ontology, such as WordNet, has been widely used for semantic search.

Lexical search is not adequate to represent the semantics of queries referring to word senses, for instance: (1) Search for documents about "*movement*"; (2) Search for documents about "*movement belonging to change*"; and (3) Search for documents about "*movement belonging to the act of changing location from one place to another*". That is because the word *movement* has many different senses. In fact, the first query searches for documents containing not only the word *movement* but also words that are its synonyms, e.g. *motion*, *front*, *campaign*, and *trend*, or hypernyms, e.g. *change*, *occurrence*, *social group*, *venture*, and *disposition*. For the second query, users do not expect to receive answer documents about words that are also labelled "*movement*", e.g. *movement belonging to a natural event* and *movement belonging to a venture*, but are not changes. Meanwhile, the third query requests documents about a precisely identified word sense. The word *movement* means not only the action of changing something but also the act of changing location from one place to another, e.g. *the movement of people from the farms to the cities*.

To choose the intended sense of a word in a context, a Word Sense Disambiguation (WSD) algorithm is employed. Supervised WSD systems have high accuracy ([19]) but need manually sense-tagged corpora for training. In IR, training corpora of a supervised WSD algorithm need to be large, which are usually laborious and expensive to create. Knowledge-based WSD systems ([20], [18], [1]) were developed to overcome the knowledge acquisition bottleneck and avoid manual effort. Besides, for specific domains, knowledge-based WSD systems have better performance than generic supervised WSD systems trained on balanced corpora ([2]).

Traditional knowledge-based WSD algorithms typically rank concepts that represent senses of a word and assign the sense with the highest rank to the word. If there are more one sense with the same highest rank, the systems will randomly choose one of those senses or choose all those senses. That is also applied to WordNet-based IR systems using WordNet as the knowledge base for WSD ([22], [12], [8], [25], [24], [6], [16]). However, that may choose a wrong sense, whence many irrelevant documents may be retrieved.

In this paper, we propose a new WordNet-based IR model in which a word is represented by its most specific meaning as possibly determined in a context. That is, in a context, after a disambiguation procedure, if a word has only a sense with the highest rank, then the word will be represented by that sense. Otherwise, if a word has more than one sense with the equally highest rank, then the most specific common hypernym of those senses will be chosen and the word will be represented by the pair of that hypernym and the form of the word.

The remainder of the paper is organized as follows. In the next section, we survey and classify WordNet-based IR research works. Section 3 describes the proposed system architecture and methods to annotate and expand queries and documents. Section 4 presents evaluation of the proposed model and comparison to other models. Finally, section 5 gives some concluding remarks and suggests future works.

## 2. A SURVEY AND CLASSIFICATION OF WORDNET-BASED SEMANTIC SEARCH APPROACHES
### 2.1 WordNet Words
WordNet ([17], [5]) is a lexical database for English organized in synonym sets (synsets). WordNet is reputed as a lexical ontology. Its version 2.1 contains about 150,000 words organized in over 115,000 synsets. Each WordNet word (WW) in a text may be annotated with its form (f), a direct hypernym of its sense, or its sense if existing in the ontology of discourse. That is, a fully recognized WordNet word has three features, namely, form, direct hypernym, and sense. For instance, in the text in Figure 2.1, a possible full annotation of the WordNet word *apple* at (1) and (2) is the WW triple (*apple*, *edible_fruit*_1, *#07739125-noun*) while for the *apple* at (3) and (4) it is the WW triple (*apple*, *apple_tree*_1, *#12633994-noun*), where *Malus pumila* and *apple* are synonyms of the same WW sense whose identifier is #12633994-*noun*.

> "To determine if an apple (1) is ready to be picked, place a cupped hand under the fruit, lift and gently twist. If the apple (2) doesn't come away easily in your hand, then it's not ready to harvest."[1]
>
> "A round, firm fruit with juicy flesh; the tree bearing this fruit, Malus pumila, comes from the family Rosaceae (rose family). There are many, many types of apples (3) grown all over the world today and these can be divided into eating, cooking and cider apples (4)."[2]

**Fig. 2.1.** Text passages from the BBC[3]:

However, due to ambiguity in a context, performance of a WSD algorithm, or limitation of the ontology of discourse, an ontology word may not be fully annotated or may have multiple annotations. As shown in Figure 2.2, the first, third, and eleventh senses among more than 11 senses of the word "*movement*" have the common hypernym *change*_3. So, with a context such as "*movement belonging to change*", the word "*movement*" can be not fully annotated (*movement*, *change*_3, *#\**) or have three annotations, i.e. (*movement*, *change*_3, *#movement*_1-*noun*), (*movement*, *change*_3, *#movement*_2-*noun*) and (*movement*, *change*_3, *#movement*_3-*noun*).

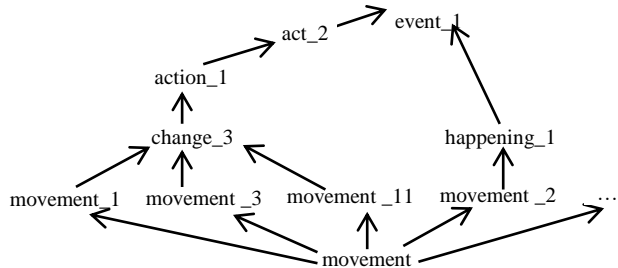

**Fig. 2.2.** Example about senses of the word "*movement*"

In this paper, we introduce the notion of most specific common hypernyms. The most specific common hypernym is a semantic relation between a sense and sense set, denoted by msc_hypernym. A sense *s* is said to be a most specific common hypernym of a sense set {$s_1$, $s_2$, ...} if *s* is one common hypernym of the sense set and no common hypernym of the sense set is more specific than *s*. For example, *event*_1 is a msc_hypernym of the four senses *movement*_1, *movement*_2, *movement*_3, and *movement*_11. We note that a sense set may have more than one msc_hypernym. Besides, we write *possible_senses*(*f*) to denote possible senses of the form *f* in a certain context.

In addition, *f/msc_hypernym*(*possible_senses*(*f*)) is combined information of a form and its respective msc_hypernym. We propose to use this combined information when a word has more than one sense determined by the WSD algorithm. For example, in a context, the WSD algorithm may determine the four possible senses *movement*_1, *movement*_2, *movement*_3, and *movement*_11 for the word "*movement*". Then, the word "*movement*" is presented by *movement/event*_1.

In summary, annotation of an ontology word having word form *f* can be one of the following formats: (1) word sense (*s*), when the sense of the word is determined; (2) combined information *f/msc_hypernym*(*possible_senses*(*f*)), when the word has more than one determined sense. The synonyms, hypernyms, and hyponyms of an ontology word can be derived from its sense.

## 2.2 WordNet based Information Retrieval Systems

With general limitations of lexical search, semantic search is a major topic of current IR research. A semantic search method often embeds semantic information in queries and/or documents and may expand them with related information. Ontology is widely used in semantic search. WordNet is the ontology of interest in this paper. Depending on the purpose and structure of an employed ontology, IR systems will use appropriate methods to exploit the ontology. Therefore, in this paper we only survey research works using WordNet for query or document semantic annotation and expansion.

Query expansion is the process of adding to an original query new terms that are similar to the original words in the query to improve retrieval performance ([14]). The works [22], [12] and [8] expanded queries by using WordNet. Document expansion is the process of enriching documents by related terms to improve retrieval performance. The works [25], [24], [6] and [16] expanded documents by using WordNet.

Table 2.1 presents our survey about text IR systems using WordNet features in comparison to our proposed one. In that, we use the following notations: (1) *s* is sense of a word; (2) *form*(*s*) is any form of a sense *s*; (3) *hypernym*(*s*) is any hypernym of a sense *s*; (4) *hyponym*(*s*) is any hyponym of a sense *s*; (5) *f/msc_hypernym*(*possible_senses*(*f*)) is the pair of a form *f* and its respective msc_hypernym in a certain context; and (6) keyword is a word that is not a stop-word or a WW.

As shown in the table, [22] expanded a query with all forms of every sense *s* occurring in it. Also, [22], [12] and [8] used all forms of a sense and all forms of any hyponym of a sense in a query. Meanwhile, [25] used all forms of a sense to expand a document, and [24] and [6] additionally used all forms of any hypernym of a sense in a document. The work [16] used senses in both queries and documents, and all forms of any hypernym of a sense in a document.

In [22], the authors showed that the use of synonyms, hyponyms and their combination in queries derived from description statements improved retrieval performance, but reduced retrieval performance with queries derived from narrative statements. In [12], after the sense of a word in a query was determined, its synonyms, hyponyms, definition were considered by some rules to be added into the query. Meanwhile, [8] expanded a query by using spreading activation on all relations in WordNet and selecting only the words being important and found in WordNet to represent the query content.

---

[1] http://www.bbc.co.uk/gardening/basics/techniques/growfruitandveg_harvestapples1.shtml

[2] http://www.bbc.co.uk/dna/h2g2/A12745785

[3] http://www.bbc.co.uk

**Table. 2.1.** Survey about search engines using features of WordNet

| Paper | | Lucene | [22] | [22], [12], [8] | [25] | [24], [6] | [16] | Our Search |
|---|---|---|---|---|---|---|---|---|
| **IR Model** | | Lexical Search | QE_Syn | QE_Syn_Hypo | DE_Syn | DE_Syn_Hyper | DE_Id_Hyper | DE_MscHyper |
| **Feature** | The features are used in | | | | | | | |
| s | Query | | | | | | x | x |
| | Doc | | | | | | x | x |
| f/msc_hypernym(*possible_senses*(f)) | Query | | | | | | | x |
| | Doc | | | | | | | x |
| form(s)/hypernym(s) | Query | | | | | | | x |
| | Doc | | | | | | | x |
| form(hyponym(s)) | Query | | | | x | | | |
| form(hypernym(s)) | Doc | | | | | x | x | x |
| hyponym(s) | Query | | | | | | | |
| hypernym(s) | Doc | | | | | | | x |
| form(s) | Query | | x | x | | | | |
| | Doc | | | | x | x | | x |
| keyword | Query | x | x | x | x | x | x | x |
| | Doc | x | x | x | x | x | x | x |

Addition to [25], [6] used hypernyms and rules to filter senses determined by the employed WSD algorithm to expand documents. If there were still more than one suitable sense for a word after running the WSD algorithm and filtering, all of those senses were used. In [25] and [6], the authors constructed concepts using the format Lemma-POS-SN, where Lemma was a form, POS was the part of speech, and SN was the sense number of the word. For example, if *surface* is a noun and has sense 1, then it will be denoted by *Surface-Noun*-1.

In [24], the authors modified the index term weights that were normally computed based on the tf*idf scheme. In a document, the weight of an index term will be increased when the term had semantic relations with other co-occurring terms in the document. The authors used WordNet to determine semantic relations between the terms. In [16], each WW was replaced by the new format Sense|POS. For example, if *surface* is a noun and has offset 3447223, then it will be denoted by *3447223|Noun*. The authors also used forms of hypernyms of a word in WordNet to increase performance of the system.

Like all systems expanding queries, the above query expansion systems spend time for searching related terms in an ontology and matching between the new query and documents. Meanwhile, in document expansion systems, searching for related terms is offline and the query is not changed. Hence, our search applies document expansion.

Moreover, since the above-surveyed papers use word forms to represent word senses, it may reduce the precision of system. Indeed, a query containing a word having form *f* and sense *x* could also match to documents containing a word having the same form *f* but different sense *y*. For example, with query expansion, for the query "*search documents about rainfall*", *rain* as a synonym of *rainfall* is added into the query. Therefore, documents about "*a rain of bullets*" will also be retrieved. Similarly, with document expansion, for the document about "*rainfall*", *rain* as a synonym of *rainfall* is added into the document. Therefore, the document will be retrieved by the query "*search documents about a rain of bullets*". The drawback is similar with using only word forms of hypernyms and hyponyms of senses.

Especially, in case a word has more than one sense determined by a WSD algorithm, the above works choose randomly one sense from those senses, which may decrease the retrieval performance if that is a wrong choice. In contrast, in our system, such a word is represented by the combination of its form and the msc_hypernym of the senses.

Besides, if a word *w* in a document has only one suitable sense *s*, then all forms, hypernyms, and form-hypernym pairs of *s* are virtually added into the document. Otherwise, if it is uttered by a form *f* in a document and determined by the pair *f/msc_hypernym*(*possible_senses*(*f*)), then *f*, *msc_hypernym*(*possible_senses*(*f*)), and all hypernyms of *msc_hypernym*(*possible_senses*(*f*)) are virtually added into the document.

## 3. THE PROPOSED WORDNET-BASED SEMANTIC SEARCH

### 3.1 System Architecture

Our proposed system architecture is shown in Figure 3.1. The WordNet Word Disambiguation-and-Annotation module extracts and embeds the most specific WW features in a raw query and a raw text document. The process is presented in detail in Sections 3.2 and 3.2. After that, the text is indexed by contained WW features and keywords, and stored in the Extended WordNetWord-Keyword Annotated Text Repository. Semantic document search is performed via the WordNet Word-Keyword-Based Generalized VSM (Vector Space Model) module as presented in Section 3.4.

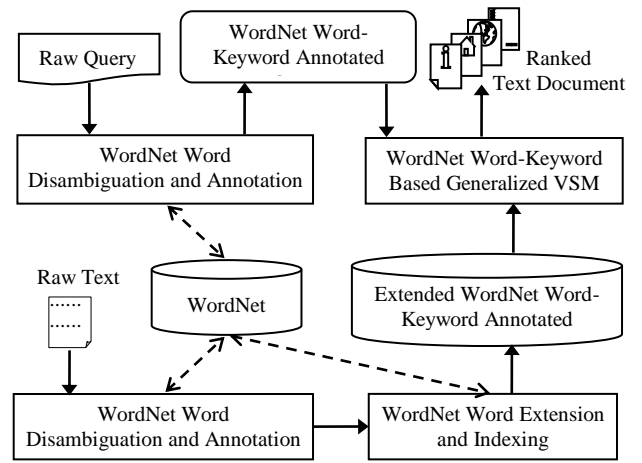

**Fig. 3.1.** System architecture for WordNet based semantic search

### 3.2 Word Sense Disambiguation using WordNet

Word sense disambiguation is to identify the right meaning of a word in its occurrence context. Lesk's algorithm ([11]) was one of the first WSD algorithms for phrases. The main idea of Lesk's algorithm was to disambiguate word senses by finding the overlap among their sense definitions using a traditional dictionary. The works [13] and [4] proposed to use WordNet for Lesk's algorithm. Following [13], we modify Lesk's algorithm by exploiting associated information of each sense of a word in WordNet, including its definition, synonyms, hyponyms, and hypernyms. By comparing the associated information of each sense of a word with its surrounding words, we can identify its right sense. However, if a word has two or more suitable senses, then our WSD algorithm will find msc_hypernyms of the senses in hypernym hierarchy of WordNet. We use WordNet version 2.1 for the WSD algorithm. Figure 3.2 describes the difference between the traditional KB-based WSDs and our KB-based WSD.

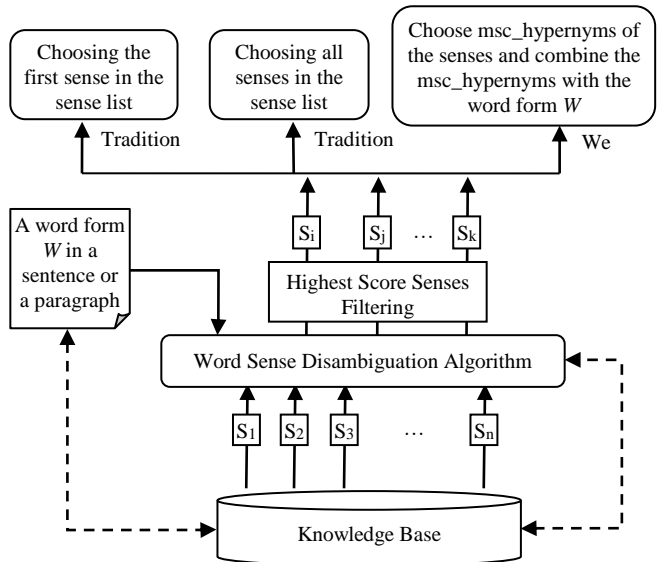

**Fig. 3.2:** Difference between the traditional KB-based WSDs and our KB-based WSD

### 3.3 Query and Document Annotation and Expansion

The system processes queries for WordNet-based searching in the following steps:
1. Removing stop-words in the query.
2. Disambiguating and annotating WordNet words in the query. We semi-automatically carry out this task for the experiments.
3. Representing each recognized WordNet word:
   - If the sense *s* of the word is determined, then the word is represented by *s*.
   - If the word has more than one sense with *f* and *msc_hypernym*(*possible_senses*(*f*)) as its apparent form and the most specific common hypernym, respectively, then the word is represented by *f*/*msc_hypernym*(*possible_senses*(*f*)).
4. Words not defined in WordNet are treated as plain keywords.

A document is automatically processed in the following steps:
1. Removing stop-words in the documents by using a built-in function in Lucene[4], which is a general open source for storing, indexing and searching documents [7].
2. Disambiguating and annotating WordNet words in the document by using our WSD algorithm introduced in above sections.
3. Extending the document with implied information:
   - If the sense *s* of the word is determined, then *s* and its expanded features *form*(*s*), *hypernym*(*s*), *form*(*hypernym*(*s*)), *form*(*s*)/*hypernym*(*s*) are added to the document.
   - If the word has more than one sense with *f* and *msc_hypernym*(*possible_senses*(*f*)) as its apparent form and the most specific common hypernym, respectively, then *f* and *f*/*msc_hypernym*(*possible_senses*(*f*)) and their expanded features

     *form*(*msc_hypernym*(*possible_senses*(*f*))),

     *msc_hypernym*(*possible_senses*(*f*)),

     *form*(*hypernym*(*msc_hypernym*(*possible_senses*(*f*)))

     *hypernym*(*msc_hypernym*(*possible_senses*(*f*))),

     *f*/*hypernym*(*msc_hypernym*(*possible_senses*(*f*)))

     are added to the document.
4. Words not defined in WordNet are treated as plain keywords.
5. Original WordNet features, implied WordNet features and plain keywords are indexed.

### 3.4 Vector Space Model with Combined WordNet Words and Plain KeyWords

In many cases, there are some words that are plain keywords or not updated yet in the exploited word ontology. We propose a VSM that combines WordNet words and plain keywords. That is, we unify and treat all of them as generalized terms, where a term is counted either as a WordNet word or a plain keyword or but not both. Each document or query is then represented by a single vector over that generalized term space. the weights of the generalized term are calculated as the in traditional *tf.idf* scheme.

We implement the above VSM by employing and modifying Lucene. In fact, Lucene uses the traditional VSM with a tweak on the document magnitude in the cosine similarity formula for a query and a document. But this does not affect the ranking of the documents. In Lucene, a term is a character string and term occurrence frequency is computed by exact string matching, after keyword stemming and stop-word removal. Here are our modifications of Lucene for what we call *S-Lucene* for the above VSM: (1) indexing documents over the generalized term space; (2) modifying Lucene codes to compute dimensional weights for the vectors representing a document or a query; and (3) modifying Lucene codes to compute the similarity degree between a document and a query.

## 4. EXPERIMENTS
### 4.1 Dataset and Performance Measures

Evaluation of a retrieval model or method requires two components being a test dataset and quality measures ([3], [15]).

---

[4] http://lucene.apache.org

The L.A. Times document collection is employed, which was used by 15 papers among the 33 full-papers of SIGIR-2007 and SIGIR-2008 about text IR using TREC dataset. The L.A. Times consists of more than 130,000 documents in nearly 500MB. Next, queries in the Adhoc Track-1999, which has answer documents in this document collection, is used. So, there are 44 queries of 50 queries in this Track chosen. Each query has three portions, namely, the title, description and narrative ones. Since a query title is short and looks like a typical user query, we only use query titles in all experiments, as in [22], [12], [8], [25], and [6], for instance.

We have evaluated and compared the IR models in terms of precision-recall (P-R) curves, F-measure-recall (F-R) curves, and single mean average precision (MAP) values ([3], [10], [15]). Meanwhile, MAP is a single measure of retrieval quality across recall levels and considered as a standard measure in the TREC community ([23]).

Obtained values of the measures presented above might occur by chance. Therefore, a statistical significance test is required ([9]). We use Fisher's randomization (permutation) test for evaluating the significance of the observed difference between two systems, as recommendation of [21]. As shown [21], 100,000 permutations were acceptable for a randomization test and the threshold 0.05 of the two-sided significance level, or two-sided p-value, could detect significance.

## 4.2 Testing results

We present experiments about search performance of our system in comparison with the surveyed WordNet–based systems by seven different search models:

1. Lexical Search: This search uses Lucene text search engine as a tweak of the traditional keyword-based VSM.
2. QE_Syn: The search uses synonyms of WordNet to expand queries only.
3. QE_Syn_Hypo: The search is similar to QE_Syn but it uses both synonyms and forms of hyponyms to expand queries.
4. DE_Syn: The search uses synonyms of WordNet to expand documents. It employs the traditional KB WordNet-based WSD as presented in Section 3.2.
5. DE_Syn_Hyper: The search is similar to DE_Syn but it uses both synonyms and forms of hypernyms to expand documents.
6. DE_Id_Hyper: The search uses the sense of a word to represent the word and forms of hypernyms of the sense to expand documents.
7. DE_MscHyper: This search uses our proposed model and system presented in Section 3.

In QE_Syn and QE_Syn_Hypo query expansion models, the sense of a word in a query is semi-automatically determined to get high precision. In DE_Syn, DE_Syn_Hyper and DE_Id_Hyper document expansion models, Lesk's algorithm is modified to automatically determine the sense of a word as for our DE_MscHyper model. However, when a word in a context has many suitable senses, the other document expansion models will choose the first ranked sense in the senses to represent the word.

Table 4.1 and Figure 4.1 plots present the average precisions and F-measures of Lexical Search, QE_Syn, QE_Syn_Hypo, DE_Syn, DE_Syn_Hyper, DE_Id_Hyper and DE_MscHyper models at each of the standard recall levels. It shows that DE_MscHyper performs better than the other six models, in terms of the precision and F measures. The MAP values in Table 4.2 and the two-sided p-values in Table 4.3 show that taking into account latent ontological features in queries and documents does enhance text retrieval performance. In terms of the MAP measure, Semantic Search performs about 17.7% better than the Lexical Search model, and about 37% and 127.1% better than the QE_Syn, and QE_Syn_Hypo models, and 9.4%, 19.2% and 23.2% better than the DE_Syn, DE_Syn_Hyper and DE_Id_Hyper, respectively.

**Table 4.1.** The average precisions and F-measures at the eleven standard recall levels on 44 queries of the L.A. Times

| Measure | Model | Recall (%) | | | | | | | | | | |
|---|---|---|---|---|---|---|---|---|---|---|---|---|
| | | 0 | 10 | 20 | 30 | 40 | 50 | 60 | 70 | 80 | 90 | 100 |
| Precision (%) | Lexical Search | 51 | 42 | 34 | 28 | 25 | 22 | 16 | 13 | 10 | 9 | 8 |
| | QE_Syn | 46 | 35 | 30 | 23 | 21 | 18 | 15 | 13 | 8 | 7 | 6 |
| | QE_Syn_Hypo | 34 | 22 | 20 | 15 | 14 | 12 | 8 | 5 | 3 | 2 | 2 |
| | DE_Syn | 51 | 45 | 37 | 30 | 27 | 23 | 18 | 14 | 12 | 10 | 10 |
| | DE_Syn_Hyper | 46 | 40 | 34 | 29 | 25 | 22 | 16 | 13 | 11 | 10 | 9 |
| | DE_Id_Hyper | 51 | 40 | 35 | 26 | 22 | 19 | 13 | 11 | 10 | 8 | 8 |
| | DE_MscHyper | 62 | 49 | 41 | 33 | 28 | 24 | 18 | 15 | 11 | 10 | 9 |
| F-measure (%) | Lexical Search | 0 | 13 | 19 | 20 | 21 | 22 | 18 | 16 | 12 | 11 | 10 |
| | QE_Syn | 0 | 12 | 17 | 18 | 19 | 19 | 17 | 15 | 11 | 9 | 9 |
| | QE_Syn_Hypo | 0 | 9 | 13 | 13 | 13 | 13 | 10 | 8 | 6 | 4 | 4 |
| | DE_Syn | 0 | 13 | 20 | 22 | 24 | 23 | 20 | 17 | 15 | 13 | 12 |
| | DE_Syn_Hyper | 0 | 13 | 19 | 22 | 22 | 22 | 18 | 16 | 14 | 12 | 11 |
| | DE_Id_Hyper | 0 | 13 | 19 | 21 | 20 | 19 | 14 | 13 | 11 | 10 | 10 |
| | DE_MscHyper | 0 | 14 | 21 | 24 | 24 | 23 | 20 | 17 | 14 | 12 | 11 |

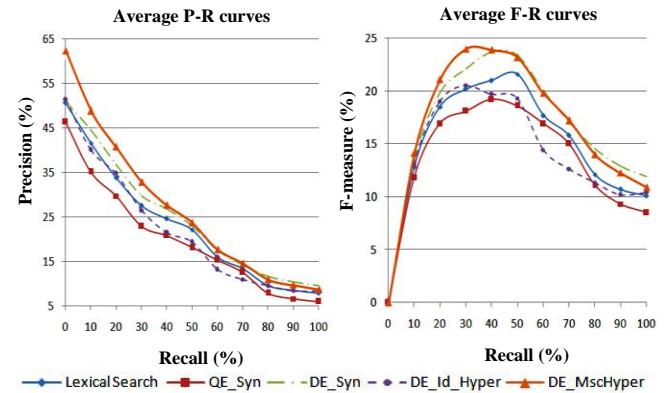

**Fig. 4.1.** Average P-R and F-R curves of Lexical Search, QE_Syn, DE_Syn, DE_Id_Hyper and DE_MscHyper models on 44 queries of TREC

**Table 4.2.** The mean average precisions on the 44 queries of TREC

| Model | DE_Msc Hyper | Lexical Search | QE_Syn | QE_Syn_ Hypo | DE_Syn | DE_Syn _Hyper | DE_Id_ Hyper |
|---|---|---|---|---|---|---|---|
| **MAP** | **0.251** | 0.2133 | 0.1832 | 0.1105 | 0.2295 | 0.2106 | 0.2037 |
| **Improvement** | | 17.7% | 37% | 127.1% | 9.4% | 19.2% | 23.2% |

**Table 4.3.** Randomization tests of DE_MscHyper with the Lexical Search, QE_Syn, QE_Syn_Hypo, DE_Syn, DE_Syn_Hyper and DE_Id_Hyper models

| Model *A* | Model *B* | \|MAP(*A*) – MAP(*B*)\| | N⁻ | N⁺ | Two-Sided P-Value |
|---|---|---|---|---|---|
| DE_MscHyper | Lexical Search | 0.0377 | 1335 | 1453 | 0.02788 |
| | QE_Syn | 0.0678 | 1181 | 1195 | 0.02376 |
| | QE_Syn_Hypo | 0.1405 | 161 | 153 | 0.00314 |
| | DE_Syn | 0.0215 | 11878 | 11574 | 0.23452 |
| | DE_Syn_Hyper | 0.0404 | 3763 | 3826 | 0.07589 |
| | DE_Id_Hyper | 0.0473 | 2187 | 2268 | 0.04455 |

## 5. CONCLUSION AND FUTURE WORKS

We have presented a generalized VSM that exploits all ontological features of WordNet words for semantic text search. That is a whole IR process, from a natural language query to a set of ranked answer documents. The conducted experiments on a TREC dataset have shown that our WordNet features exploitation improves the search quality in terms of the precision, recall, F, and MAP measures.

For future works, we are considering combination of WordNet and other ontologies to increase the number of WordNet words that can be covered. Also, we are researching rules to determine WordNet words that are not updated into employed ontologies.